\documentclass[%
 reprint,
 amsmath,amssymb,
 aps,
 prx,
floatfix,
]{revtex4-2}

\usepackage{graphicx}% Include figure files
\usepackage{dcolumn}% Align table columns on decimal point
\usepackage{bm}% bold math
\usepackage[ruled,linesnumbered,vlined]{algorithm2e}
\usepackage{multirow}
\usepackage{enumitem}
\usepackage{hyperref}
\usepackage{cleveref}
\begin{document}

\preprint{APS/123-QED}

\title{Variational Functional Neural Networks (vfNN)}% Force line breaks with \\

\author{Chinmay Katke$^{1, 2}$}
\author{C. Nadir Kaplan$^{1-4}$}%
 \email{nadirkaplan@vt.edu}
\affiliation{%
 $^1$Department of Physics,
Virginia Polytechnic Institute and State University, Blacksburg, VA 24061, USA. 
\\$^2$Center for Soft Matter and Biological Physics,
Virginia Polytechnic Institute and State University, Blacksburg, VA 24061, USA. 
\\$^3$Department of Mechanical Engineering,
Virginia Polytechnic Institute and State University, Blacksburg, VA 24061, USA. 
\\$^4$Center for the Mathematics of Biosystems,
Virginia Polytechnic Institute and State University, Blacksburg, VA 24061, USA.
}%

\date{\today}% It is always \today, today,

\begin{abstract}
We present a variational functional neural network (vfNN) algorithm that optimizes physically motivated continuum functionals in integral form (e.g., Lagrangian, Hamiltonian, or free energy) to find the states of a system. That way, vfNN bypasses solving the equations of motion (a.k.a. the Euler-Lagrange equations), which are in the form of differential equations. To this end, our approach is fundamentally different from conventional physics-informed neural networks (PINN) that solve differential equations by minimizing a phenomenological loss function involving higher order derivatives and ad hoc weight factors. In contrast, vfNN merely evaluates lower order derivatives of the generating functional and replaces the weight factors with rigorous physical scales. This also fundamentally changes the loss landscape, and the desired solutions may become unstable. For this emergent nonconvex optimization problem, we develop a quasi-Newton optimization algorithm with simulated annealing that converges to the solutions of the corresponding Euler-Lagrange equations of the functional regardless of their stability properties. Our simulations using vfNN give accurate results for benchmark systems governed by linear or nonlinear ordinary differential equations.
\end{abstract}

\maketitle

%\tableofcontents

\section{Introduction}
The equations of motion of a wide range of dissipative or energy conserving physical systems can be derived from a Lagrangian, Hamiltonian, or Rayleighian functional that involves conservative or nonconservative forces. The analytical extremization yields Euler-Lagrange equations (ELE), which inherently employ gradient information to solve for a smooth function, such as the position of a body in time, or an order parameter and a nonuniform vector field of a many-body system.

Neural networks as universal function approximators and reliable gradient algorithms have thus become appealing for automating the solution of physically motivated differential equations~\cite{hornik1989multilayer, rall1996introduction}. To this end, physics-informed neural networks (PINN) have become a popular tool to solve for the state of a system by minimizing a phenomenological loss function constructed from a set of coupled differential equations~\cite{RAISSI2019686, Toscano2025}. This loss function is the $\mathcal{L}^2$ norm of the set, and the derivative terms in the loss function are calculated using automatic differentiation (AD)~\cite{rall1996introduction}. The procedure is mesh-free in the sense that the accuracy of the solution does not depend on the coarseness of the underlying mesh. This gives PINN a significant advantage over conventional differential equation solvers in problems involving irregular domains and large deformations~\cite{Zhang2022multi, Hu2024solving}. Consequently, PINN found wide applicability in biomedicine, fluid and solid mechanics, materials science, geophysics, dynamical systems and control, and chemical engineering~\cite{Daryakenari2024, Yazdani2020, JO2024100899, Qian2024-fz, JIN2021109951, Eivazi2022, JAGTAP2022111402, Florio2021, Li_2021, Bastek_2023, Waheed_2021, Nazari2024, Wu2023}. 

Despite the advantages of NN implementations in physical modeling, using AD to compute higher order derivatives, which lie at the heart of the mathematical description of physical systems, is expensive due to extensive mapping of all lower derivatives. Even if the differential equations are cast to a coupled system of first-order equations, the corresponding optimization procedure using PINN involves construction of a single loss function with ad hoc weight factors multiplying each equation. These weight factors are estimated manually or algorithmically, increasing the complexity of the problem. 

These shortcomings can be circumvented by turning to the variational formulation of the problem at hand, which naturally has lower order derivatives. A variational PINN framework that utilizes weak formulation has previously been implemented in the Petrov-Galerkin framework (VPINN), where a deep neural network (DNN) acts as a trial function, and the test functions are represented by prescribed functions or a second neural network \cite{kharazmi2019variationalphysicsinformedneuralnetworks, KHARAZMI2021113547, Berrone2022, zang2020weak, khodayi-mehr20a}. These formulations are similar to the finite element method (FEM), and the accuracy depends on the number of quadrature points in the domain and the choice of test functions. Another NN approach -- much closer in spirit to the physically motivated variation as with Lagrangian and Hamiltonian mechanics -- was explored in the context of the Poisson's equation as an elliptic partial differential equation that was numerically solved by directly minimizing a convex functional~\cite{E2018, LIU2023116229, URIARTE2023115892, lu2021priorigeneralizationanalysisdeep, SAMANIEGO2020112790, NGUYENTHANH2021114096, KARUMURI2020109120, nguyen2020deep, SHENG2021110085}. 

Here, we generalize this latter approach by developing a variational functional neural network (vfNN) algorithm to directly optimize a broader class of physically motivated functionals that have nonconvex loss landscapes. This approach bypasses solving ELE, thus significantly differing from PINN, as well as from VPINN in that the variational form is not the weak form. We demonstrate that vfNN can successfully extremize a functional to find the solution(s) even when the loss landscape is not guaranteed to be locally or globally convex. We implement vfNN to a range of benchmark problems with a variational formulation, whose ELE would correspond to a single (linear or nonlinear) ordinary differential equation (ODE) or a coupled set thereof. 

%We characterize the stability of the solutions to ELE using the notion of conjugate points and we present an algorithm to find conjugate points for linear ordinary differential equations. 

This paper is organized as follows: in Sec.~\ref{sec:methods}, we present the details of the vfNN algorithm. In particular, we characterize the nonconvexity of the loss landscape to assess the stability of the solutions of ELE, and then develop a quasi-Newton optimization procedure that identifies the solutions regardless of their stability. In Sec.~\ref{sec:results}, we utilize vfNN to extremize the functionals of different physical systems whose ELEs constitute a linear ODE, a nonlinear ODE for the minimal surface of revolution, steady-state Sine-Gordon equation, Poisson-Boltzmann equation, and a set of coupled ODEs for the packing of hard rods in a liquid crystalline colloidal monolayer. We provide a discussion and concluding remarks in Sec.~\ref{vfPINN_conclusion}.

\section{Methods}
\label{sec:methods}

As a basic variational problem, we seek a function $\mathbf{\hat{u}(x)}$ that optimizes the integral in a domain $\Omega \subset \mathbb{R}^d$

\begin{alignat}{4}
    &\mathcal{J}& = &  \int \limits_{\Omega} F(\mathbf{x}, \mathbf{u(x)}, \mathbf{u'(x)}) \,d\mathbf{x}\,, \quad \mathbf{x}=(x_1, x_2,...,x_d)\in \Omega\,, \label{Eq:basic var problem}\\  
    &&&\mathbf{u}=  (u_1, u_2,...,u_n)\,, \qquad  u'_{kj}\equiv\frac{\partial u_k}{\partial x_j}\,.\nonumber
\end{alignat}
Here $F$ is a scalar function of unknowns $(u_1, u_2,..,u_n)$ and their gradients $(\mathbf{u'_1}, \mathbf{u'_2},.., \mathbf{u'_n})\,.$ The optimization is subject to the following Dirichlet ($D$) and Neumann ($Nm$) boundary conditions:
\begin{equation}
\begin{split}
    \mathbf{u(x)}&=\mathbf{u}_{D}\,, \hspace{1cm} \mathbf{x}\in \partial\Omega_{D}\,,\\
    \mathbf{u'(x)} &= \mathbf{u}_{Nm}\,, \hspace{1cm} \mathbf{x}\in\partial\Omega_{Nm}\,.\label{Eq:BC general problem}
\end{split}
\end{equation}
The Dirichlet and Neumann boundary conditions hold on different parts of the domain boundary $\partial\Omega_{D}$ and $\partial\Omega_{Nm}\,,$ respectively, where known functions $\mathbf{u}_{D}$ and $\mathbf{u}_{Nm}$ are imposed.

%The problem can be stated more compactly as: Given $S\,,$ the set of smooth admissible comparison functions that satisfy the boundary conditions Eq. \ref{Eq:BC general problem},

%\begin{equation}
%    \underset{u\in S}{\text{opt}} \{ \mathcal{J} | \mathbf{u(x)}=\mathbf{u_D}, \hspace{0.25cm} x\in\partial\Omega_D; \mathbf{u'(x)}=0, \hspace{0.25cm} x\in\partial\Omega_N\}\,. \label{Eq: compact}
%\end{equation}

By considering first-order variation of Eq.~\ref{Eq:basic var problem} with respect to $\mathbf{u}\,,$ we arrive at the corresponding Euler-Lagrange equations (ELE): 

\begin{equation}
    \frac{\partial F}{\partial u_k} = \frac{d}{dx_j} \frac{\partial F}{ \partial u'_{kj}}\,. \label{Eq: euler-lagrange}
\end{equation}
The solution $\mathbf{\hat{u}(x)}$ to the ELE is defined as the first-order critical point in variational calculus \cite{gelfand2000calculus}. 

%(what are the other ways to solve variational problem and why use neural networks - introduction)

To solve Eq.~\ref{Eq:basic var problem}, we represent $\mathbf{u(x)}$ using a neural network (NN) as a universal function approximator, which reduces the original infinite-dimensional optimization problem to finite dimensions~\cite{hornik1989multilayer}. A point $\mathbf{x}_i\in\Omega$ is an input to the neural network, and the corresponding output is $\mathbf{u_N}(\mathbf{x}_i)$. We denote the set of such points in the domain by $\{\mathbf{x}_i\}$ (collocation points) and the set of the outputs from the neural network by $\{\mathbf{u_N}(\mathbf{x}_i)\}\,.$ 

\begin{figure}
    \centering
    \includegraphics[width=1.0\linewidth]{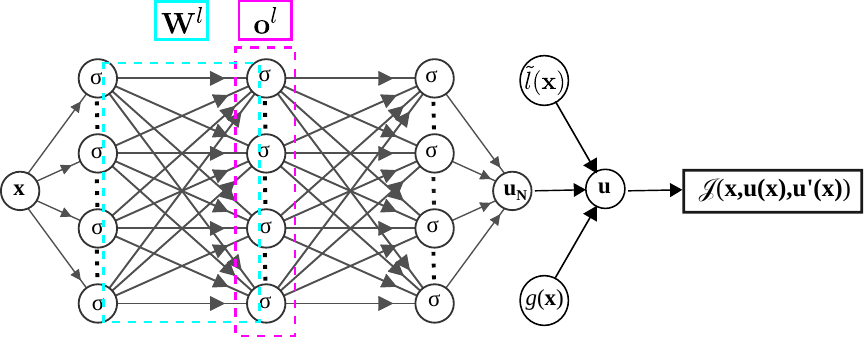}
    \caption{\textbf{Variational functional neural network (vfNN)}. We use a multilayer perceptron (MLP) deep neural network with an input $\mathbf{x}$ and output $\mathbf{u}$ that minimizes the variational loss $\mathcal{J}\,.$ The derivative $\mathbf{u'(x)}$ is computed using automatic differentiation at each input point $\mathbf{x}\,.$ The Dirichlet boundary conditions are implemented by transforming the output of the network $\mathbf{u_N}$ with known functions $\tilde{l}(\mathbf{x})$ and $g(\mathbf{x})\,.$ The Neumann boundary conditions are implemented by including a corresponding boundary term in the functional $\mathcal{J}\,.$ For the definitions of $\sigma, l, \mathbf{W}, \mathbf{o}\,,$ see Eqs.~\ref{eq:preactivation},~\ref{eq:activation}.}
    \label{fig:fig 1 vfPINN architecture}
\end{figure}

The network topology that we use for vfNN is depicted in Fig.~\ref{fig:fig 1 vfPINN architecture}. We construct the NN as follows: let $\mathbf{o}^0 \equiv \mathbf{x}_i$ denote the input to the network, and let $m$ be the number of layers. For each layer $l=1,\dots,m$ with the number of neurons $n_l\,,$ let $\mathbf{W}^l \in \mathbb{R}^{n_l \times n_{l-1}}$ be the weight matrix, and $\mathbf{b}^l \in \mathbb{R}^{n_l}$ be the bias vector that enables nonzero output of the neural network for zero input ($\mathbf{x}=0$). The output of the $(l-1)$\textsuperscript{th} layer is denoted by $\mathbf{o}^{l-1}\,.$ Then,
\begin{align}
  \mathbf{a}^l &= \mathbf{W}^l\cdot\mathbf{o}^{l-1} + \mathbf{b}^l\,, \quad l=1,2,\dots,m\,, \label{eq:preactivation} \\
  \mathbf{o}^l &= \sigma^l\!\left( \mathbf{a}^l \right)\,, \quad l=1,2,\dots,m\,. \label{eq:activation}
\end{align}
The activation functions are defined as
\begin{align}
  \sigma^l(x) &= \tanh(x)\,, \quad l=1,2,\dots, m-1\,, \label{eq:tanh} \\
  \sigma^m(x) &= x\,, \quad \text{(identity function)}\,. \label{eq:identity}
\end{align}
 We choose the hyperbolic tangent function as an activation function for all the layers except for the last layer, which allows for a simple pass-through to give the output $o^m=\mathbf{u_N(x_i)}\,.$ For convenience, we denote the set of all parameters that are the entries of the matrices $\mathbf{W}^l$ and vectors $\mathbf{b}^l$ by a single vector $\mathbf{w}\in\mathbb{R}^{n_t}\,,$ where ${n_t}$ is the total number of NN parameters. We choose the most fundamental neural network, multilayer perceptron (MLP), as the underlying network structure. Furthermore, we set three hidden layers (length) with 20 neurons in each layer (width). These network dimensions are optimal for the validation problems considered in this work. Networks with number of layers less than three and a width less than 20 were not sufficient to approximate the solutions. We found that MLP with bigger length and width higher increased computational cost without significantly improving the accuracy.

In the conventional formulation of PINN, the ELE is solved by minimizing the corresponding $\mathcal{L}^2$ norm~\cite{RAISSI2019686}:
\begin{equation}
    \mathcal{L}_{PINN} = \sum_i\mu_i\sum_j \left( \frac{\partial F}{\partial u_i} - \frac{d}{dx_j} \frac{\partial F}{ \partial_j u_{i}'}\right)^2\,.% \bigg |_{x=x_i}\,.
    \label{eq:L2norm}
\end{equation}
Here, the sum over index $i$ runs over the Euler-Lagrange equations, each corresponding to an unknown $u_i$ describing the system, and $\mu_i$ are the corresponding loss weight factors. The determination of these weight factors is not trivial and often involves trial and error or implementation of complex algorithms. In vfNN, to obtain the solution of Eq.~\ref{Eq: euler-lagrange}, we optimize Eq.~\ref{Eq:basic var problem} where rigorous physical scales describing the system effectively replace the loss weight factors $\mu_i$ in Eq.~\ref{eq:L2norm}. To this end, we first approximate the integral in Eq.~\ref{Eq:basic var problem} by a right Riemann sum: 
\begin{equation}
    \mathcal{J}\approx\frac{V}{N}\underset{i}{\sum}F(\mathbf{x_i}, \mathbf{u(x_i)}, \mathbf{u'(x_i)})\,.
    \label{eq:Riemann}
\end{equation}
Here $V=\idotsint_V  \,dx_1 \dots dx_d$ is the volume of the domain $\Omega\,,$ and $N$ is total number of collocation points. For a fixed set of collocation points $\{\mathbf{x}_i\}\,,$ $\mathcal{J}$ is a function of $\mathbf{w}\,,$ i.e., the variational problem in Eq.~\ref{Eq:basic var problem} can be solved by optimizing the NN weights and biases.

The most popular method to implement the boundary conditions (BCs) in PINN is to include a corresponding penalty term for the deviation from the desired BC in the overall loss function~\cite{RAISSI2019686}. For the Dirichlet BCs stated in problem \ref{Eq:basic var problem}, the penalty terms are given by 
\begin{equation}
    \mathcal{L}_{BC} = \mu\underset{i}{\sum}(\mathbf{u_N(x_i)}-\mathbf{u}_D)^2\,, \hspace{1cm} \mathbf{x}_i\in\Omega_D\,,
    \label{eq:PINN_BC}
\end{equation}
 where $\mu$ is again a penalty coefficient. Generally, the optimization is initiated with a smaller value $\mu$ whose value is increased gradually until the BCs are met. Since this method adds an extra layer of complexity to the optimization process, in vfNN, we instead implement the Dirichlet BCs for the problem in Eq.~\ref{Eq:basic var problem} by directly constraining the output of the neural network~\cite{Lu2021}:
\begin{equation}
    \mathbf{u(x)} = \tilde{l}(\mathbf{x})\mathbf{u_N}(\mathbf{x}) + g(\mathbf{x})\,. 
    \label{eq:ch5_bc}
\end{equation}
The functions $\tilde{l}(\mathbf{x})$ and $g(\mathbf{x})$ are any smooth function over the domain $\mathbf{x}\in\Omega$ with following properties:

\begin{alignat}{4}
    \tilde{l}(\mathbf{x}_i) &\neq 0\,, & \mathbf{x}_i&\in \Omega\,, \\
    \tilde{l}(\mathbf{x}_i) &= 0\,, & \mathbf{x}_i&\in \Omega_D \nonumber\,,\\
    g(\mathbf{x}_i) &= \mathbf{u}_D\,, \quad & \mathbf{x}_i&\in \Omega_D\,. \nonumber 
\end{alignat}
The simplest function one can construct to satisfy the above criteria are $\tilde{l}(\mathbf{x})$ quadratic in $\mathbf{x}$ and $g(\mathbf{x})$ linear in $\mathbf{x}$. Hence, we have chosen $\tilde{l}(\mathbf{x})$ and $g(\mathbf{x})$ accordingly for each problem studied in this work.

%The important difference between the loss functions of PINN and vfNN to be noted is that PINN seeks to minimize the loss locally at every point $\mathbf{x_i}\,,$ while in vfNN, optimum of $\mathcal{J}$ does not necessarily imply local optimization as long as global sum $\mathcal{J}$ is optimized. The functional $\mathcal{J}$ naturally involves lower order derivatives compared to $\mathcal{L}_{PINN}$ which reduces the cost associated with computation of higher order derivatives. 

Eq.~\ref{eq:L2norm} ensures that the solution is a local minimum in the multidimensional loss landscape, and the optimization algorithms such as the adaptive moment estimation (ADAM) and the limited memory Broyden–Fletcher–Goldfarb–Shanno algorithm (L-BFGS) work very well to find the solution~\cite{kingma2017adammethodstochasticoptimization, liu1989limited}. In contrast, although the solution $\mathbf{\hat{u}}$ to Eq.~\ref{Eq: euler-lagrange} is the optimum solution to the problem defined by Eq.~\ref{Eq:basic var problem}, it is not guaranteed that the optimum corresponds to a minimum (See Appendix~\ref{App:A} for the conditions of stability). In general, the solution $\mathbf{\hat{u}}$ could be located at a maximum, minimum, or a saddle point in the landscape of $\mathcal{J}\,;$ this increases the complexity of optimization. To overcome this, Appendix~\ref{App:B} outlines a simple algorithm to determine the stability of a 1D functional whose ELE is a linear ODE.

To find the solution $\mathbf{\hat{u}}$ of the general variational problem in Eq.~\ref{eq:Riemann}, we develop an optimization algorithm that converges to a first-order critical point based on Newton's method of optimization. Newton's method uses the first derivatives as well as the second derivatives or the Hessian matrix of the objective function to find the optimum (minimum, maximum or a saddle point), offering faster convergence compared to other first-order methods such as the gradient descent algorithm. However, it can suffer from higher computational and memory costs associated with the second derivatives in a large parameter space, in our case of dimension $n_t$ of the weight vector $\mathbf{w}$. The quasi-Newton second-order methods like the Broyden–Fletcher–Goldfarb–Shanno (BFGS) algorithm approximate the Hessian of the objective function using first derivatives and the step size in the parameter space, thereby reducing the computational cost. The second-order optimization methods are typically combined with supplementary algorithms to accelerate convergence. While quasi-Newton methods determine the search direction in the parameter space, the supplementary ``line search" algorithms determine the optimum step size in that direction. A variety of line search algorithms satisfy criteria called the Wolfe conditions that impose sufficient decrease (Armijo) and the curvature conditions, ensuring convergence to a global minimum when combined with the BFGS algorithm~\cite{nocedal1999numerical}. 

The conventional BFGS method approximates the true Hessian matrix by using another matrix with a strictly positive eigenvalue spectrum. In the vicinity of a local minimum, the Hessian is always positive definite (i.e., has only positive eigenvalues), and hence BFGS works very well at locating minima. While this method is perfectly suited for the task of finding a minimum, it is ineffective for finding a solution to variational problems when the solution lies at a saddle point or a maximum in the loss landscape. This is because around the saddle points, the true Hessian has both positive and negative eigenvalues. 

Thus, we use Symmetric Rank 1 (SR1) Hessian update with a line search method based on a Metropolis criterion~\cite{chib1995understanding, nocedal1999numerical}. The advantage of using SR1 Hessian update is that the approximate Hessian is not constrained to represent only the positive eigenvalues, accurately approximating the true Hessian matrix when it is indefinite (i.e., has positive, zero, and negative eigenvalues) in the vicinity of a saddle point or inflection point solution. 

%While SR1 method has been used to find the minimum of the loss function in conjunction with neural networks architecture, here we have used it to find the optimum (minimum, maximum or saddle point) of the given loss function. 

% we use the Wolfe conditions~\cite{nocedal1999numerical}

% \noindent \textbf{1. Sufficient decrease (Armijo) condition:}
% \begin{equation} \label{eq:armijo}
%   \mathcal{J}(\mathbf{w}_k + \alpha_k \mathbf{p}_k) \leq \mathcal{J}(\mathbf{w}_k) + c_1 \, \alpha_k \, \nabla \mathcal{J}(\mathbf{w}_k)^T \mathbf{p}_k\,,
% \end{equation}
% where \( 0 < c_1 < 1 \) is a small constant (typically, \(c_1 \approx 10^{-4}\)) ensuring a sufficient reduction in the objective function for a chosen step size $\alpha_k$.

% \noindent \textbf{2. Curvature condition:}
% \begin{equation} \label{eq:curvature}
%     \|\nabla \mathcal{J}(\mathbf{w}_k + \alpha_k \mathbf{p}_k)^T \mathbf{p}_k\|  \leq c_2 \, \|\nabla \mathcal{J}(\mathbf{w}_k)^T \mathbf{p}_k\| \,,
% \end{equation}
% with \( c_1 < c_2 < 1 \) (a common choice is \(c_2 \approx 0.9\)). This condition ensures that the slope has sufficiently been reduced, implying that the step \(\alpha_k\) is not too small.

We implement the line search method based on a Metropolis criterion as follows: let $\mathbf{w}_k=(w_1,..., w_n)$ be a vector where the elements $w_i$ denote the NN weights and biases, and the subscript $k$ denotes the $k$\textsuperscript{th} iteration. Let $\mathcal{J}(\mathbf{w_k})\equiv \mathcal{J}_k$ be the variational loss function at the iteration $k$, $\nabla \mathcal{J}_k$ be the gradient in the loss landscape calculated using backpropagation, and $B_k$ is an $n\times n$ matrix that approximates the Hessian matrix of $\mathcal{J}_k\,.$ The matrix $B_k$ is evaluated iteratively during optimization. We define the vector $\mathbf{p}\in\mathbb{R}^{n_t}$ as the direction of the next iteration step, i.e., the line search direction. Then, the quadratic approximation $m_k (\mathbf{p})$ of the variational loss about $\mathcal{J}_k$ is given by
\begin{equation}
    m_k(\mathbf{p}) \equiv  \mathcal{J}(\mathbf{w_k+\mathbf{p}}) =\mathcal{J}_k + \nabla \mathcal{J}_k^T\,\mathbf{p} + \frac{1}{2}\mathbf{p}^TB_k\mathbf{p}\,.
    \label{eq:ch5_quad_apprx}
\end{equation}
By also expanding $\nabla m_k(\mathbf{p})$ into a Taylor series, the optimum of the quadratic approximation in Eq.~\ref{eq:ch5_quad_apprx} is obtained by setting $\nabla m_k(\mathbf{p})=0$:
\begin{equation}
    \mathbf{p}_k = -B_k^{-1} \nabla \mathcal{J}_k+\mathcal{O}(\mathbf{p}^2)\,. \label{Eq: line search direction}
\end{equation}
Eq.~\ref{Eq: line search direction} is used to calculate the next iteration,
\begin{equation}
    \mathbf{w}_{k+1} = \mathbf{w}_k + \alpha_k \mathbf{p}_k\,,
\end{equation}
where $\alpha_k$ is the step size. 

%To determine an appropriate value for $\alpha_k$, we employ 

\begin{algorithm}
    \caption{Stochastic line search}
    \label{alg:line_search}
    \KwIn{$\mathbf{w}_k$, $\nabla\mathcal{J}(\mathbf{w}_k)$, $B_k^{-1}$, $\gamma$, $\beta$, $\alpha_{\min}$}
$\alpha \leftarrow 1$,\ $\mathbf{p}_k \leftarrow -B_k^{-1}\nabla\mathcal{J}(\mathbf{w}_k)$,\ $i \leftarrow 0$\;
\Repeat{$i = i_{\max}$}{
    $\phi(\alpha) \leftarrow \|\nabla\mathcal{J}(\mathbf{w}_k + \alpha\mathbf{p}_k)\|$\;
    \eIf{$\textit{uniform}(0,1) \leq \exp\!\left(-\beta[\Phi(\alpha)-\Phi(0)]\right)$}{
        \Return{$\alpha$}\;
    }{
        $\alpha \leftarrow 0.9\,\alpha$\;
        $i \leftarrow i+1$\;
    }
}
\Return{$\alpha_{\min}$}\;
\end{algorithm}

To find an appropriate step size $\alpha_k\,,$ our line search method given in Algorithm~\ref{alg:line_search} employs a Metropolis criterion. The resulting value of $\alpha_k$ reduces the magnitude of the objective function gradient along the line search direction $\mathbf{p}_k$ towards $\|\nabla \mathcal{J}_k\|\approx 0$ where the optimum of $\mathcal{J}$ is located. While typical line search algorithms may utilize a variety of deterministic criteria to find an optimal step size $\alpha_k$, such as the Wolfe conditions that impose the sufficient decrease (Armijo) and the curvature conditions~\cite{nocedal1999numerical}, our algorithm is stochastic and is thus much less sensitive to noise in a potentially rough and nonconvex loss landscape. At every iteration $k$, we start with the maximum allowed step size $\alpha\equiv \alpha_k=1$ and inverse ``temperature" $\beta$ for a Newton update. The Metropolis criterion
\begin{equation}
    \textit{uniform}(0,1) \leq e^{-\beta[\Phi(\alpha)-\Phi(0)]}\,, \quad \Phi(\alpha) \equiv \frac{\phi(\alpha)}{\phi(0)^{-\gamma}}\,,\label{Eq. Metropoli}
\end{equation}
will always be true if $\phi(\alpha)=\|\nabla \mathcal{J}(\mathbf{w}_k+\alpha\mathbf{p}_k)\|$ is smaller than $\phi(0)\,,$ and the step length $\alpha_k$ will be accepted. Here, $uniform(0,1)$ corresponds to a random number between $0$ and $1\,,$ sampled from the uniform distribution at each iteration, which leads to occasional acceptance of a step length with $\phi(\alpha)>\phi(0)\,.$ This helps navigate the highly nonconvex loss landscape of the neural network to find the optimum. We rescale the cost function $\phi(\alpha)$ with $\phi(0)^{-\gamma}$ to ensure that the algorithm does not randomly escape the basin of the optimum; $\gamma > 0$ is a tuning parameter. While lower values of $\gamma$ increase the chances of escaping the basin of the optimum, higher values of $\gamma$ result in very slow convergence. Thus, we choose $\gamma=0.5\,.$ If the Metropolis criterion is not met, then the step length $\alpha$ is reduced by a factor ($=0.9$ here), and the process is repeated. The failure to satisfy the Metropolis condition for the maximum number of iterations would return a small constant step length $\alpha_{min}\,.$

Having determined the line search direction $\mathbf{p}_k$ (Eq.~\ref{Eq: line search direction}) and the step size $\alpha_k$ (Algorithm~\ref{alg:line_search}), the SR1 Hessian update that relates the approximate Hessian matrices $B_{k}$ and $B_{k+1}$ is given as~\cite{nocedal1999numerical}
\begin{eqnarray}
    B_{k+1} &=& B_k + \frac{(\mathbf{y}_k - B_k\mathbf{s}_k)(\mathbf{y}_k-B_k\mathbf{s}_k)^T}{(\mathbf{y}_k - B_k\mathbf{s}_k)^T\mathbf{s}_k} \label{Eq: SR1 update} \\
    &\equiv& SR1(B_k,\mathbf{s}_k,\mathbf{y}_k)\,,\nonumber
\end{eqnarray}
with
\begin{equation}
    \mathbf{s}_k = \mathbf{w_{k+1}}-\mathbf{w_k}=\alpha_k\mathbf{p}_k\,, \quad  \mathbf{y}_k = \nabla \mathcal{J}_{k+1} -\nabla \mathcal{J}_k\,. 
\end{equation}
For the SR1 update scheme, we have defined the function $SR1(B_k\,,\mathbf{s}_k\,,\mathbf{y}_k)\,.$ Eq. \ref{Eq: SR1 update} can be obtained by considering the secant equation
\begin{equation}
    B_{k+1} \mathbf{s_k} = \mathbf{y_k}\,.
    \label{eq:secant}
\end{equation}
The only symmetric rank-1 update satisfying Eq.~\ref{eq:secant} and $B_{k+1}=B_{k}+c \mathbf{v}\mathbf{v}^\top$ ($c\in\mathbb{R}\,, \mathbf{v}\in\mathbb{R}^{n}\,,$ $n\in\mathbb{R}$) is given by Eq.~\ref{Eq: SR1 update}~\cite{nocedal1999numerical}. We start with $B_0=\mathbf{I}$, where $\mathbf{I}$ is the identity matrix. The inversion of an $n\times n$ matrix $B_k$ required to solve Eq.~\ref{Eq: line search direction} is impractical when $n\gg1$. The elegance of the quasi-Newton methods is that the inverse $B_k^{-1}$ can also be computed iteratively~\cite{nocedal1999numerical}:
\begin{eqnarray}
    B_{k+1}^{-1}=SR1(B_k^{-1},\mathbf{s}_k,\mathbf{y}_k)\,. \label{Eq: SR1 inverse update}
\end{eqnarray}
% &=& B_k^{-1} + \frac{(\mathbf{s}_k - B_k^{-1}\mathbf{y}_k)(\mathbf{s}_k-B_k^{-1}\mathbf{y}_k)^T}{(\mathbf{s}_k - B_k^{-1}\mathbf{y}_k)^T\mathbf{y}_k} \\
The SR1 update becomes unstable when the vectors $\mathbf{y}_k - B_k\mathbf{s}_k$ and $\mathbf{s}_k$ are perpendicular to each other and their scalar product in the denominator in Eq.~\ref{Eq: SR1 update} vanishes. To avoid this numerical instability, we perform the update when the criterion below is met:
\begin{equation}
    |\mathbf{s}_k^T(\mathbf{y}_k-B_k\mathbf{s}_k)|\geq r\|\mathbf{s}_k\| \|\mathbf{y}_k-B_k\mathbf{s}_k\|\,,
\end{equation}
where $r\in(0,1)$ is a very small number. If the above condition is not met, then $B_{k+1} = B_k$~\cite{nocedal1999numerical}. 

%The SR1 update has been used for optimization, specifically for the task of finding a minimum, paired up with trust-region methods \cite{Byrd1996, Erway03052020, Yousefi2023, Indra2022}. The line search methods based on Wolfe conditions are incompatible with SR1 update since they cannot accommodate indefinite Hessian. Trust-region methods look for sufficient decrease in the objective function inside an n-dimensional hyper-sphere around the current iterate $w_k$ rather than along a specific line search direction and only require $B_k$ to be symmetric. Still, trust-region methods are unsuitable for the task at hand of finding unstable first order critical points since they actively look for only the minimizers of the objective function.

\begin{algorithm}
\caption{SR1 Hessian update with simulated annealing}
\label{alg:SR1 quenching}
\KwIn{$\mathbf{w}$: weights, $\mathcal{J}(\mathbf{w})$: loss functional, $\nabla\mathcal{J}(\mathbf{w})$: gradient, $\delta$: tolerance, $\tau$: quenching rate, $T_0$: initial temperature, $T_f$: final temperature}
$B^{-1} \leftarrow \mathbf{I}$,\ $i \leftarrow 0$\;
\Repeat{$\|\nabla\mathcal{J}\| < \delta$}{
    $\beta(i) \leftarrow \dfrac{1}{(T_0 - T_f)\,\exp(-\tau i) + T_f}$\;
    $\alpha \leftarrow \text{Line search}(\beta)$\;
    $\mathbf{w}_{i+1} \leftarrow \mathbf{w}_i + \alpha\, B^{-1} \nabla\mathcal{J}(\mathbf{w}_i)$\;
    $\mathbf{s}_i \leftarrow \mathbf{w}_{i+1} - \mathbf{w}_i$,\quad $\mathbf{y}_i \leftarrow \nabla\mathcal{J}(\mathbf{w}_{i+1}) - \nabla\mathcal{J}(\mathbf{w}_i)$\;
    $B^{-1} \leftarrow \mathrm{SR1}(B^{-1},\, \mathbf{s}_i,\, \mathbf{y}_i)$\;
    $i \leftarrow i + 1$\;
}
\Return{$\mathbf{w}_{i+1}$}
\end{algorithm}

The procedure for the SR1 Hessian update with quenching is given in Algorithm~\ref{alg:SR1 quenching}, which employs simulated annealing with the inverse temperature $\beta$ varied through the optimization process~\cite{Kirkpatrick_simulated_annealing,  bertsimas1993simulated}. We start with low values of $\beta$ and quench the system by increasing $\beta$. The quenching rate $\tau$ is an additional tuning parameter. A lower quenching rate increases the likelihood of finding the optimum at a cost of higher number of iterations. The numerical values used for the parameters $T_0\,, T_f\,, \tau\,,\delta$ in Algorithm~\ref{alg:SR1 quenching} can be found in Ref.~\cite{NN_GitHub}.

\section{Results}
\label{sec:results}
By using our vfNN algorithm where the vector of NN parameters $\mathbf{w}_k$ is optimized through Algorithm~\ref{alg:SR1 quenching} over $k$ iterations, we solve several representative linear and nonlinear ordinary differential equations (ODEs) by using their variational formulation. In Figs.~\ref{fig:2 nonlinear Eq1} and~\ref{fig:2 nonlinear Eq2}, we compare our results from vfNN with FEM results that we generated by using FEniCS 2019.1.0~\cite{fenics2015}.

\begin{figure}
    \centering
    \includegraphics[width=1.0\linewidth]{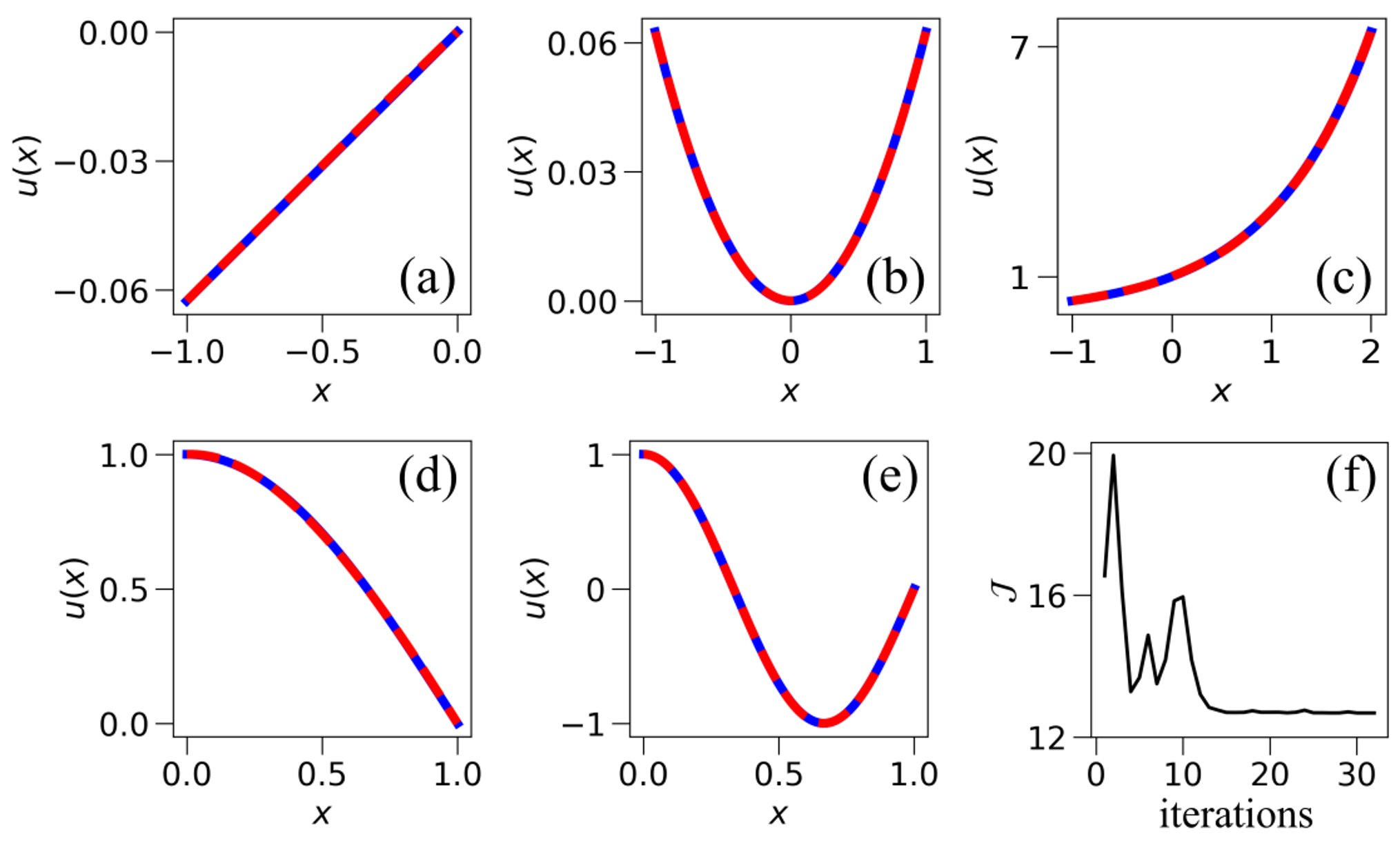}
    \caption{\textbf{Linear ODEs using vfNN.} The function $f(u)$ in Eq.~\ref{Eq: ELE Poisson} is chosen as \textbf{(a)} $f(u)=0\,,$ \textbf{(b)} $f(u)=0.25\,,$ \textbf{(c)} $f(u)=u\,,$ \textbf{(d)} $f(u)=-(\pi/2)^2u\,,$ and \textbf{(e)} $f(u)=-(3\pi/2)^2u^\ast\,.$ \textbf{(f)} shows the variational loss $\mathcal{J}$ for $f(u)=u$ in (c) as a function of the iterations of our SR1 update (Algorithm~\ref{alg:SR1 quenching}). The dashed red lines show the solutions obtained using vfNN, and the solid blue lines indicate the solutions obtained using PINN. $\ast$ The domain in (e) contains a conjugate point that makes the loss landscape nonconvex; see Appendices~\ref{App:A},~\ref{App:B}.}
    \label{fig:1 Poisson equation}
\end{figure}

\begin{figure}
    \centering
    \includegraphics[width=0.8\columnwidth]{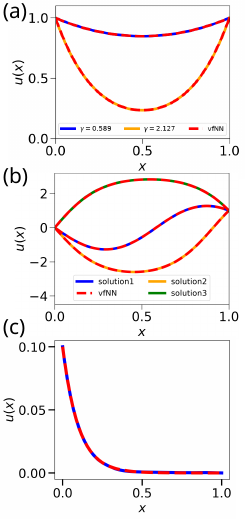}
    \caption{\textbf{Nonlinear ODEs using vfNN.} Solutions to the functionals whose ELE are \textbf{(a)} the minimal surface of revolution (Eqs.~\ref{eq:minimal_functional},~\ref{eq:minimal_exact}), \textbf{(b)} steady-state Sine-Gordon equation (Eq.~\ref{eq:sine-Gordon}), and \textbf{(c)} the Poisson-Boltzmann equation (Eqs.~\ref{eq:PB_functional}--\ref{eq:PB_Dirichlet}). The dashed red lines indicate the solutions using vfNN, and all other colors correspond to different nonunique solutions (a) from the exact solution (Eq.~\ref{eq:minimal_exact}) and (b), (c) found by using FEM. For all panels, a deep neural network with three layers and 20 neurons at each layer is used.}
    \label{fig:2 nonlinear Eq1}
\end{figure}

\begin{figure}
    \centering
    \includegraphics[width=0.8\columnwidth]{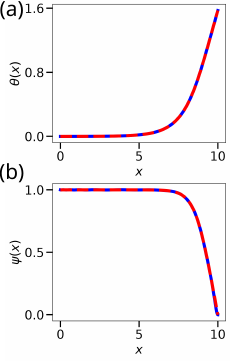}
    \caption{\textbf{Nonlinear ODEs using vfNN (continued).} Minimization of the elastic energy of a colloidal membrane composed of hard rods (Eqs.~\ref{eq:hard_rod_energy}--\ref{eq:hard_rod_BC})~\cite{Kaplan_har_rods_2014}. \textbf{(a)} $\theta(x)$ the tilt angle with respect to membrane normal. \textbf{(b)} $\psi(x)$ is the scalar order parameter describing the orientation of the rods with respect to the membrane normal.  The $x$ axis is nondimensionalized by the twist penetration depth $\lambda_t\,.$ The membrane interior is at $x=0\,,$ whereas $x=L=10$ is the position of the membrane edge. The dashed red lines indicate the solutions using vfNN, and the blue curves correspond to the solutions from FEM. Two separate deep neural networks (three layers with 20 neurons at each layer) for each of the coupled dependent variables $\theta(x)$, $\psi(x)$ were used. }
    \label{fig:2 nonlinear Eq2}
\end{figure}

We first consider the variational formulation of a linear second order ODE
\begin{equation}
     \mathcal{J}=  \int_{a}^{b} F(x, u(x), u'(x)) \,dx\,, \hspace{1cm} F = u'(x)^2 + f(u)u\,, \label{Eq: variational problem Poisson}
\end{equation}
whose ELE is
\begin{equation}
    \frac{d^2u}{dx^2} = \frac{1}{2}\left(\frac{\partial f}{\partial u}u + f\right)\,. \label{Eq: ELE Poisson}
\end{equation}
Here, $f(u)$ represents a known function up to linear order in $u\,.$ Table~\ref{tab:fu_boundary_conditions} lists the Dirichlet boundary conditions that we implement at both endpoints of the respective domains for different choices of $f(u)\,.$  The results for the choices of $f(u)$ listed in Table~\ref{tab:fu_boundary_conditions} are depicted in Fig.~\ref{fig:1 Poisson equation}a--e. Unlike conventional PINN where the loss function is an $\mathcal{L}^2$ norm (Eq.~\ref{eq:L2norm}), the loss function of vfNN may converge to a nonzero value. One such example is $f(u)=u\,,$ and Fig.~\ref{fig:1 Poisson equation}f shows the convergence of the corresponding variational loss $\mathcal{J}$ as a function of iteration steps. 
\begin{table}[htbp]
  \centering
  \caption{\textbf{Functions \( f(u) \) and corresponding boundary conditions for Eqs.~\ref{Eq: variational problem Poisson},~\ref{Eq: ELE Poisson}.} The results are shown in Fig.~\ref{fig:1 Poisson equation}.}
  \begin{tabular}{|c|c|}
    \hline
    \( f(u) \) & Boundary conditions \\
    \hline
    \(0\) & \(u(-1.0) = -0.06\,,\quad u(0) = 0\,.\) \\
    \(0.25\) & \(u(-1) = 0.06\,,\quad u(1.0) = 0.06\,.\) \\
    \(u\) & \(u(-1) = e^{-1}\,,\quad u(2) = e^2\,.\) \\
    \(-\left(\frac{\pi}{2}\right)^2\)$u$ & \(u(0) = 1\,,\quad u(1) = 0\,.\) \\
    \(-\left(\frac{3\pi}{2}\right)^2\)$u$ & \(u(0) = 1\,,\quad u(1) = 0\,.\) \\
    \hline
  \end{tabular}
  \label{tab:fu_boundary_conditions}
\end{table}

We also compare the convergence rates of vfNN and PINN, each subject to the loss optimization through our SR1 algorithm and the conventional ADAM algorithm~\cite{kingma2017adammethodstochasticoptimization}. For each of the network-algorithm pair, Table \ref{Table:comparison vfPINN and PINN} presents the number of iterations necessary to achieve accurate solutions. The iterations are averaged over 30 independent initializations due to the randomness associated with our line search algorithm. Except for the case of $f(u)=(3\pi/2)^2u\,,$ vfNN along with SR1 algorithm for loss optimization requires least number of iterations for convergence. This computational advantage can be attributed to two factors; the reduced computational cost due to lower order derivatives in the loss function $\mathcal{J}$ and the superlinear convergence rate of the quasi-Newton SR1 optimization algorithm. Note that, for $f(u)=-(3\pi/2)^2u\,,$ the use of ADAM optimizer on vfNN  results in a divergence of the loss. In this case, the sole solution to the ELE resides at an unstable saddle point within the loss landscape, and the ADAM optimizer, which seeks the minimum of the loss function, is unable to locate the solution. Here, the first two sufficiency conditions for a stable solution given in Appendix~\ref{App:A} are satisfied since $\hat{F}_{u',u'}=1\,$. Thus, it must be the third condition that is violated, in other words, the presence of the conjugate points makes the solution unstable. We find the sole conjugate point $x_c=0.66668$ in the domain using our algorithm~\ref{Alg:findingconjpoints} (Appendix~\ref{App:B}) in 14 iterations which is accurate up to 4\textsuperscript{th} decimal. This algorithm also finds multiple conjugate points, $\{0.66668,1.33334\}\,,$ in the extended domain $x\in[0,1.8]\,.$ Thus, we employ the SR1 optimizer to accurately identify the corresponding unstable solution in the domain $x\in[0,1]$ without the need to identify the conjugate points.

\begin{table}
\centering
    \begin{tabular}{|p{1.5cm}|c|c|c|c|}
       \hline
       \multirow{2}{1.5cm}{$f(u)$} & \multicolumn{2}{c|}{vfNN} & \multicolumn{2}{c|}{PINN}\\
       \cline{2-5}
        & SR1 & ADAM & SR1 & ADAM \\ \hline
        & \multicolumn{4}{c|}{Number of iterations}\\
        \hline
        $0$ &28 & 603&120 & 442\\ \hline
        $0.25$ &28 &262&68 & 832\\ \hline
        $u$ &456 &854 &761 & 499\\ \hline
        $-\left(\frac{\pi}{2}\right)^2 u$ &5&229 &54 &413 \\ \hline
        $-\left(\frac{3\pi}{2}\right)^2 u$ &1793 & - & 1396 & 1985\\ \hline
    \end{tabular}
    \caption{\textbf{Convergence rates of vfNN vs. PINN}. By specifying different $f(u)$ for the problem given in Eq.~\ref{Eq: variational problem Poisson}, we optimize PINN and vfNN via two algorithms: the first-order optimization algorithm ADAM~\cite{kingma2017adammethodstochasticoptimization} and the second-order optimization algorithm SR1 (Sec.~\ref{sec:methods}). For $f(u)=-(3\pi/2)^2u\,,$ the use of ADAM optimizer with vfNN leads to the divergence of the loss function since the domain contains a conjugate point that makes the loss landscape nonconvex; see Appendices~\ref{App:A},~\ref{App:B}. For all cases, a deep neural network with three layers and 20 neurons per layer is used. The number of iterations are averaged over 30 independent initializations.}
    \label{Table:comparison vfPINN and PINN}
\end{table}

%Fig.~\ref{fig:2 nonlinear Eq1} shows the results for the implementation of vfNN to solve nonlinear ODE. 

We next consider the functionals of a single dependent variable and its derivative whose ELEs are a single nonlinear ODE. First, we consider the problem of finding the minimal surface of revolution formed by the rotation of a two-dimensional curve around the $x$-axis. Since the desired function must always lie above the $x$-axis, such that it has no simple root, we impose the constraint $u(x)=[\tilde{l}(x)u_N(x)+g(x)]^2\,,$ where $u_N(x)$ is the output of the neural network and does not directly represent the dependent variable (see Fig.~\ref{fig:fig 1 vfPINN architecture}). The variational formulation of this problem is
\begin{equation}
     \mathcal{J}=  \int_{a}^{b} u\sqrt{1+u'(x)^2} \,dx\,, \quad u(x)\geq0\,.
     \label{eq:minimal_functional}
\end{equation}
We choose a domain $x \in [0,1]$ with the Dirichlet boundary conditions $u(0)=u(1)=1\,.$ The special solution of the problem can be found analytically as
\begin{equation}
    u(x) = \frac{1}{2\gamma}\text{cosh}\left[\gamma(2x-1)\right]\,.
    \label{eq:minimal_exact}
\end{equation}
The neck radius of the surface of revolution $u_{min}$ satisfies $u_{min}=1/2\gamma\,.$ There exist two solutions to this problem: while the solution corresponding to $\gamma=0.589388$ is a minimum, the solution corresponding to $\gamma=2.1268$ is a saddle point (Fig.~\ref{fig:2 nonlinear Eq1}a). Our vfNN algorithm successfully identifies both solutions by initializing the optimization from different initial guesses that are randomized by the initialization of the weight matrix $\mathbf{W}^l$ in Eq.~\ref{eq:preactivation}.

The second nonlinear problem with a single dependent variable that we consider is the steady-state Sine-Gordon equation, which is also an integrable system and models the large-angle simple pendulum when the independent variable $x$ represents time. The variational formulation and the corresponding ELE of the problem is as follows:
\begin{equation}
     \mathcal{J}=  \int_{a}^{b} \left[\frac{1}{2}u'(x)^2 + \omega^2 \cos u\right] \,dx\,,  \quad \frac{d^2u}{d x^2} = -\omega^2\sin u\,. 
     \label{eq:sine-Gordon}
\end{equation}  
In the domain $x\in [0,1]\,,$ imposing Dirichlet boundary conditions at both boundaries, $u(0)=0$ and $u(1)=1$ for $\omega=6\,,$ this problem has three distinct solutions: solution 1 and solution 2 are stable (corresponding to the equilibrium of the pendulum at $u=0$ and $u=2\pi$), and solution 3 is unstable (corresponding to $u=\pi$). By changing the initial guess, vfNN locates all three solutions regardless of their stability in agreement with the FEM profiles (Fig.~\ref{fig:2 nonlinear Eq1}b).

The third nonlinear problem with a single variable we investigate is the Poisson-Boltzmann equation. The problem herein describes the electrostatic potential $u$ in the proximity of a uniform planar charge distribution in the presence of an electrolyte solution as function of the normal distance $x$ from the charged surface. For a constant $\alpha$ determined by system properties such as the surface charge density, valency of the screening charges in the electrolyte solution, and temperature, the variational formulation corresponding to mixed boundary conditions in a domain $x\in\left[a\,,\, b\right]$ is
\begin{equation}
    \mathcal{J} = \int_{a}^{b} \left(\frac{1}{2}u'(x)^2 + \alpha^2 \cosh(u)\right) \,dx -\beta u |_{x=a}\,,
    \label{eq:PB_functional}
\end{equation}
where $\beta_E$ corresponds to the normalized electric field near the charged surface. The last term of Eq.~\ref{eq:PB_functional} indicates how a Neumann condition is applied in vfNN. The corresponding ELE and the Neumann boundary condition at $x=a$ is given by
\begin{equation}
    \frac{d^2 u}{d x^2} = \alpha^2 \sinh(u)\,, \quad \frac{d u}{d x}\bigg|_{x=a} = \beta_E\,.
    \label{eq:PB_equation}
\end{equation}
We complete the setup by choosing our parameters as $a=0\,,b=1\,,\alpha=10\,,$ $\beta_E=0.1\,,$ and impose the Dirichlet boundary condition at $x=b$,
\begin{equation}
    u|_{x=b} = 0\,.
    \label{eq:PB_Dirichlet}
\end{equation}
The results from vfNN and FEM agree very well (Fig.~\ref{fig:2 nonlinear Eq1}c), indicating that vFNN reliably tackles nonlinear problems with mixed boundary conditions.

Finally, we consider a nonlinear problem with two coupled dependent variables in a single functional that models a colloidal flat monolayer of monodisperse chiral hard rods. Whereas the rods are perfectly aligned in the normal direction to the membrane surface in the interior of the monolayer because of polymer-induced depletion interactions, the tilting of rods towards the membrane edge due to molecular chirality $q_0$ (with the units of inverse length) can be treated as a packing problem with liquid crystalline symmetry, specifically as a single layer of a chiral Smectic-A (Sm-A$^\ast$) phase that melts into a cholesteric (Ch) order toward the monolayer edge~\cite{Kaplan_har_rods_2014, zakhary2014imprintable}. The rod tilting with an angle $\theta(x)$ with respect to the monolayer surface normal manifests itself as twist deformations with a penetration depth $\lambda_t\,,$ and the melting of Sm-A$^\ast$ into Ch can be represented by a phase field $\psi(x)$ with a coherence length $\xi_{||}$ akin to the real part of the smectic order parameter. Defining a unitless parameter $\kappa\equiv \lambda_t/\xi_{||},$ the unitless form of the free energy nondimensionalized by $\lambda_t$ in a unitless domain $x\in\left[0, L\right]$ ($x=0:$ monolayer bulk $x=L:$ monolayer edge) is given by~\cite{Kaplan_har_rods_2014, zakhary2014imprintable}
%\begin{widetext}
\begin{eqnarray}
    \mathcal{J}(\psi, \theta) = \int_{0}^{L}&&dx\left[-\psi^2 + \frac{\psi^4}{2} + \frac{1}{\kappa^2}\left(\frac{d \psi}{dx}\right)^2\right.\nonumber\\  &&+ \left.\psi^2\sin^2\theta + \left(\frac{d \theta}{dx} - q_0\lambda_t\right)^2\right]\,.
    \label{eq:hard_rod_energy}
\end{eqnarray}
%\end{widetext}
The corresponding Euler-Lagrange equations for $\psi(x)$ and $\theta(x)$ are
\begin{eqnarray}
    \frac{1}{\kappa^2} \frac{d^2\psi}{d x^2} + \psi\left(1-\sin^2\theta\right) - \psi^3 &=& 0\,,\\
    \frac{d^2 \theta}{dx^2} - \frac{1}{2} \psi^2 \sin2\theta &=& 0\,.
    \label{eq:hard_rod_ELE}
\end{eqnarray}
We set the parameters as $\kappa=10\,,$ $L=10\,,$ $q_0\lambda_t=0\,,$ and impose the following Dirichlet boundary conditions:
\begin{eqnarray}
    \psi |_{x=0} &= 1\,, \hspace{1cm} \theta |_{x=0} &= 0\,,\\
    \psi |_{x=L} &= 0\,, \hspace{1cm} \theta |_{x=L} &= \frac{\pi}{2}\,.
    \label{eq:hard_rod_BC}
\end{eqnarray}
To improve the convergence rate, we formulate the order parameter in terms of the NN output as $\psi(x)=\left [\tilde{l}(x) u_N(x)+g(x)\right]^2\,,$ since the parity symmetry of $\psi$ evident in Eq.~\ref{eq:hard_rod_energy} otherwise leads to a highly degenerate loss landscape. When the rods are achiral ($q_0\lambda_t=0$), twist deformations can still be observed since the rod tilting minimizes the membrane surface exposed to the depletant polymer solution~\cite{Kaplan_har_rods_2014, zakhary2014imprintable}. In our theory, the twist of achiral rods is enforced by the Dirichlet condition at $\theta |_{x=L}$ in Eq.~\ref{eq:hard_rod_BC}. The results are shown in Fig.~\ref{fig:2 nonlinear Eq2} where the profiles from vfNN are compared to the FEM data, indicating excellent agreement for this benchmark model with more than one dependent variables, each represented by a separate neural network that has three layers and 20 neurons per layer.

\section{Discussion and Conclusions}
\label{vfPINN_conclusion}
In this paper, we have presented the variational functional neural network (vfNN) that finds the states of a system by optimizing a physically motivated variational formulation. Many physical systems are described by an integral functional $\mathcal{J}$ as formulated in Eq.~\ref{Eq:basic var problem}. The analytical extremization of $\mathcal{J}\,,$ i.e., $\delta\mathcal{J}=0$ yields the Euler-Lagrange equations, which are the equations of motion that determine the physical states. As opposed to PINN (cf. Eq.~\ref{eq:L2norm}), vfNN uses $\mathcal{J}$ as the error function and optimizes it with two clear strengths over PINN: first, vfNN employs characteristic physical scales that govern the relative weights of the terms in the error function $\mathcal{J}\,,$ instead of the ad hoc weight coefficients $\mu_i$. Second, because $\mathcal{J}$ always has lower order derivatives than the EL equations, we have found that vfNN consistently improves the computation times with respect to PINN for many functionals as fewer backpropagation steps are needed to take the lower order derivatives. 

While a variational formulation provides significant computational advantage due to a loss function with lower order derivatives, stable solutions of the variations do not always exist. To address this issue, our vfNN algorithm incorporates quasi-Newton optimization based on the SR1 Hessian update paired with a stochastic line search method based on a Metropolis criterion. Our optimizer successfully converges to the first-order critical points regardless of their stability for ODEs, highlighting its potential in addressing problems where traditional optimizers, such as ADAM, may fail. 

In its current state, our algorithm is agnostic to the nature of the optima, i.e., whether they are maxima, minima, saddle, or inflection points. Finding the inflection points corresponding to the ``conjugate points" (Appendices~\ref{App:A} and~\ref{App:B}) is of paramount importance because of their physical significance: depending on the problem, characterizing the conjugate points can reveal dynamical or phase behavior, as well as the universal properties of a physical system. For example, in a magnetic system where magnetization is defined by a  function $m(T)$ ($T:$ temperature), there is a ferromagnetic-to-paramagnetic transition at a critical temperature $T_c$, which is a conjugate point. At $T=T_c$, the free energy of the system, which is the error function to be optimized, has an inflection point. When $T<T_c$, the free energy landscape develops a saddle point where it becomes nonconvex. The point $T=T_c$  leads to a pitchfork bifurcation~\cite{strogatz2015nonlinear}. So, if the identification and characterization of conjugate points can be automated, vfNN could calculate the phase transitions of a many-body system or bifurcations and stability of a dynamical system. To find the conjugate points together with the physical solutions $\mathbf{\hat{u}}$ in a nonconvex landscape, an iterative algorithm that assesses local stability during optimization is needed.

Future work will further aim to extend vfNN to PDEs where the domain of independent variables is composed of the space and time dimensions. However, a generic vfNN PDE solver should treat space and time as independent variables on equal grounds. We will incorporate this approach to develop a general vfNN that can tackle the canonical forms of elliptic, parabolic, and hyperbolic linear PDEs before an extension to nonlinear PDEs are considered as an ultimate goal. 

\begin{acknowledgments}
We thank J. Barone, S. Cheng, J. Gray, M. Pleimling for fruitful discussions and the VT College of Science for financial support. We acknowledge the VT Advanced Research Computing Center for computing resources.
\end{acknowledgments}

\section*{Data Availability}
The data and codes that support the findings of this article are openly available at Ref.~\cite{NN_GitHub}.

\appendix

\counterwithin*{equation}{section}
\renewcommand\theequation{\thesection\arabic{equation}}

\section{Stability of a functional with one independent variable}\label{App:A}

To determine the stability of a functional, the second variation in $\mathcal{J}$ must be considered~\cite{gelfand2000calculus}. Here we present simplified picture using a single unknown $u(x)$ in one dimension, $x\in[a,b]\,.$ Let variation around solution $\hat{u}$ be given by $u\equiv\hat{u}+\epsilon h\,,$ where $\epsilon$ is a scalar and $h$ is smooth function which obeys boundary conditions:

\begin{equation}
\begin{split}
    h(x)&=0\,, \hspace{1cm} x\in \partial\Omega_D \,,\\
    h'(x) &= 0\,, \hspace{1cm} x\in\partial\Omega_{Nm} \,.\label{Eq:BC general problem h}
\end{split}
\end{equation}
Then, the second variation in $\mathcal{J}$ is given by,

\begin{widetext}
\begin{eqnarray}
    I[h]\equiv\frac{d^2J}{d\epsilon^2}\Bigg|_{\epsilon=0} 
    =\int_{a}^{b} \left[\hat{F}_{u,u}(x)h^2(x) + 2\hat{F}_{u,u'}h(x)h'(x) + \hat{F}_{u',u'}(x)h'^2(x)\,\right]dx\,. 
    \label{Eq:second variation}
\end{eqnarray}
\end{widetext}
For $\hat{u}$ to be a minimum, $I[h]$ must be positive for all admissible $h\,.$ This forms a secondary variational problem in $h\,.$ The corresponding Euler-Lagrange equation is called Jacobi's accessory equation \cite{gelfand2000calculus}:

\begin{eqnarray}
    [P(x)h'(x)]'&=& Q(x)h(x)\,,\quad\text{where}\nonumber\\ \quad P(x) &\equiv& \hat{F}_{u',u'}(x)\,, \nonumber\\ \quad Q(x)&\equiv& \hat{F}_{u,u}-[\hat{F}_{u,u'}(x)]\,.
    \label{Eq:Jacobi eqn}
\end{eqnarray}
Here, $\hat{F}_{u,u}\,,$ $\hat{F}_{u',u'}$ and $\hat{F}_{u,u'}$ denote the second functional derivatives with respect to their subscripts. Eq.~\ref{Eq:Jacobi eqn} is a second order differential equation in $h(\mathbf{x})$ with boundary conditions specified by Eq.~\ref{Eq:BC general problem h}. Let $\hat{h}(\mathbf{x})$ be the solution to Eq. \ref{Eq:Jacobi eqn} satisfying boundary conditions given in \ref{Eq:BC general problem h}. If
\begin{enumerate}[labelindent=0pt, noitemsep, topsep=0.0em]
    \item $P(x)>0\;,\forall x\in[a,b]$\,,
    \item $\hat{h}(\mathbf{x})$ has no roots in $(a,b]$\,,
\end{enumerate}
then the second variation $I[h]$ is positive definite; thus, $\hat{u}$ is a minimum \cite{gelfand2000calculus}. The roots of $\hat{h}(\mathbf{x})$ are defined as the conjugate points to the boundary point $a$. The three sufficiency conditions that determine if $\hat{u}$ is a weak minimum (defined as a minimum over the function space of continuous but not necessarily smooth functions) to the variational problem are 
\begin{enumerate}[labelindent=0pt, noitemsep, topsep=0.0em]
    \item $\hat{u}$ \label{minimum_criterion1}is a solution to the Euler-Lagrange equation (Eq.~\ref{Eq: euler-lagrange}),
    \item $\hat{u}$ \label{minimum_criterion2} obeys the Legendre condition, $\hat{F}_{u',u'}>0$ $\forall x\in[a,b]\,,$ 
    \item the domain of $\hat{u}$ has no conjugate points to its boundaries. 
\end{enumerate}

\section{Locating conjugate points of functionals with a single linear ordinary Euler-Lagrange equation}\label{App:B}

We have developed Algorithm~\ref{Alg:findingconjpoints} to find the conjugate points for a linear ODE by using the bisection method without solving the Jacobi's accessory equation, which requires knowing $\hat{u}$ in the first place. An example for the extremization of a functional from this algorithm is shown in Fig.~\ref{fig:1 Poisson equation}c.

This algorithm has two significant limitations: first, we assume that Legendre condition is satisfied while looking for the conjugate points. Second, it only applies to the linear ODEs since their solution is unique for the given boundary condition. Hence, the solution to the variational problem in the given domain is either a minimum or not a minimum, and this property is exploited to find the conjugate points. 

\begin{algorithm}
\caption{Finding conjugate points}
\label{Alg:findingconjpoints}
\KwIn{$\{x_{\text{start}}, x_{\text{end}}\}$: domain,\ $\mathcal{J}$: loss functional,\ $\delta_1$: tolerance,\ $\delta_2$: tolerance for convergence of vfNN,\ $N$: epochs}
$x_{\text{left}} \leftarrow x_{\text{start}}$,\ $x_{c,\text{list}} \leftarrow [\,]$\;
Optimize vfNN on $x \in [x_{\text{left}},\, x_{\text{end}}]$ using ADAM for $N$ epochs\;
\eIf{$\Delta\mathcal{J}/\mathcal{J} < \delta_2$}{
    \Return{$x_{c,\text{list}}$}\;
}{
    \Repeat{$\Delta\mathcal{J}/\mathcal{J} < \delta_2$}{
        $x_a \leftarrow x_{\text{left}}$,\ $x_b \leftarrow x_{\text{end}}$\;
        \Repeat{$x_b - x_a < \delta_1$}{
            $x_c \leftarrow (x_a + x_b)/2$\;
            Optimize vfNN on $x \in [x_{\text{left}},\, x_c]$ using ADAM for $N$ epochs\;
            \eIf{$\Delta\mathcal{J}/\mathcal{J} < \delta_2$}{
                $x_a \leftarrow x_c$\;
            }{
                $x_b \leftarrow x_c$\;
            }
        }
        Append $(x_a + x_b)/2$ to $x_{c,\text{list}}$\;
        $x_{\text{left}} \leftarrow x_c$\;
    }
    \Return{$x_{c,\text{list}}$}\;
}
\end{algorithm}

\pagebreak


\bibliography{apssamp2}% Produces the bibliography via BibTeX.

\end{document}